\documentclass[letterpaper, 10 pt, conference]{ieeeconf}  

\IEEEoverridecommandlockouts                              

\usepackage{cite}
\usepackage{amsmath,amssymb,amsfonts}
\usepackage{algorithmic}
\usepackage{graphicx}
\usepackage{textcomp}
\usepackage{xcolor}
\usepackage{verbatim} 
\usepackage{hyperref}
\usepackage{tabularx}
\usepackage{multirow}

\title{\LARGE \bf
LiDAR-based curb detection for ground truth annotation in automated driving validation 
}

\author{Jose Luis Apellániz$^{1}$, Mikel García$^{1,2}$, Nerea Aranjuelo$^{1,2}$, Javier Barandiarán$^{1}$ and Marcos Nieto$^{1}$
\thanks{$^{1}$Fundación Vicomtech, Basque Research and Technology Alliance (BRTA),
        Donostia - San Sebastián (Spain)}%
\thanks{$^{2}$Universidad del País Vasco (UPV/EHU),
         Donostia - San Sebastián (Spain)}%
}

\begin{document}

\maketitle
\thispagestyle{empty}
\pagestyle{empty}

\begin{abstract}

Curb detection is essential for environmental awareness in Automated Driving (AD), as it typically limits drivable and non-drivable areas. 
Annotated data are necessary for developing and validating an AD function.
However, the number of public datasets with annotated point cloud curbs is scarce. 
This paper presents a method for detecting 3D curbs in a sequence of point clouds captured from a LiDAR sensor, which consists of two main steps.
First, our approach detects the curbs at each scan using a segmentation deep neural network. Then, a sequence-level processing step estimates the 3D curbs in the reconstructed point cloud using the odometry of the vehicle. 
From these 3D points of the curb, we obtain polylines structured following ASAM OpenLABEL standard. These detections can be used as pre-annotations in labelling pipelines to efficiently generate curb-related ground truth data. We validate our approach through an experiment in which different human annotators were required to annotate curbs in a group of LiDAR-based sequences with and without our automatically generated pre-annotations. The results show that the manual annotation time is reduced by $50.99\%$ thanks to our detections, keeping the data quality level.
\end{abstract}


\section{Introduction}

Automated vehicles rely on different sensors to understand their near environment and make decisions accordingly. Cameras provide rich semantic information about the scene but lose depth information in their projective process, making them less aware of 3D structures than Light Detection and Ranging (LiDAR) sensors. LiDAR sensors are unaffected by light conditions \cite{maalej2018vanets}, and due to the three-dimensional nature of the point clouds generated by the sensor, they are particularly useful for the precise localization of obstacles in a vehicle's surroundings. Given the information LiDARs and cameras provide about an ego-vehicle environment, these sensors are part of many Automated Driving (AD) systems. They are often combined with the latest advances in Artificial Intelligence (AI).


The evolution and rapid improvements in AI in the last years have fueled advances in autonomous driving functions. However, validating an advanced driving function is a major challenge. Currently, the trend to evaluate these systems is to test the algorithms against extremely large volumes of accurately labelled ground truth data \cite{KALRA2016182}. Obtaining annotated data is a tedious, time-consuming, and expensive task. Large-scale public datasets might alleviate this task. 
However, their content and variability are limited, and they have task-specific annotations (e.g., cuboids for 3D object detection). For that reason, annotation tools that allow and facilitate the annotation task with semi-automatic functionalities or pre-annotations are highly demanded \cite{mujika2019web}. Machine learning algorithms can be used to provide the target information (e.g., cars). However, ground-truth data require very high accuracy and do not allow for errors, so using them with no human manual intervention is impractical in most situations.

\begin{figure*}
\centering
\includegraphics[width=\textwidth]{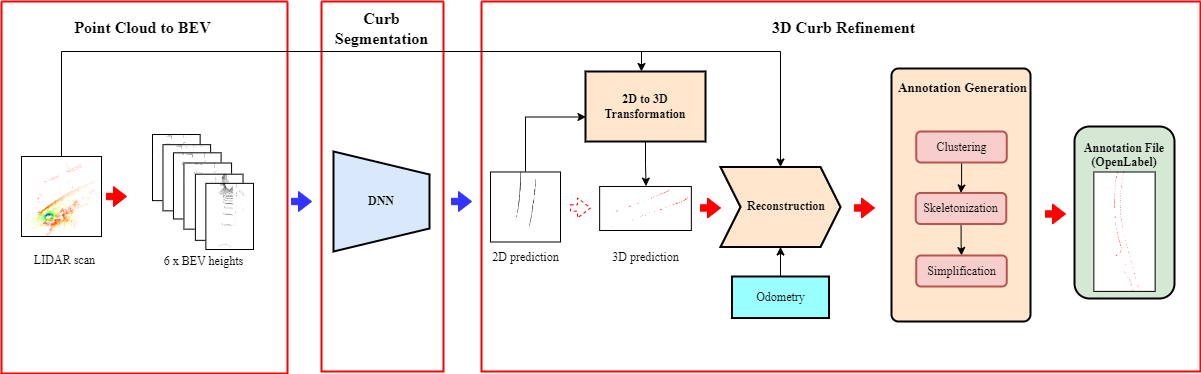}
\caption{Proposed curb annotation pipeline. A previously trained DNN takes as input a six-level bird's eye view height map to give a 2D curb prediction. Then, after transforming these 2D predictions to 3D by adding the height information from the original point cloud, a reconstruction process of all the sequences is carried out. The annotation generation post-processing is performed to get an annotation file which might be loaded into the labelling tool.}
\label{fig_diag_general_v3}
\end{figure*}

Curb detection in autonomous driving is fundamental for a complete understanding of the vehicle's environment. Curbs are part of road boundaries, separating drivable and non-drivable areas, and are an essential reference in AD tasks such as autonomous parking or route planning \cite{romero2021road}. The detection of curbs is also critical for the validation of many advanced driving functions, as they delimit potential areas of interest (e.g., parking slots and sidewalks). Curb annotated data is very scarce but necessary.


To alleviate the manual annotation task, we propose a methodology to provide 3D curbs' pre-annotations that can be incorporated in labelling tools like \cite{mujika2019web} for semi-automatic annotation. 
Our method consists of two stages. First, we perform a coarse curb detection using a deep neural network (DNN) per scan. Then, a post-processing step refines the scan-level detections and provides the curbs' pre-annotations for the whole input sequence, which can be used as input to the labelling tool and are represented as polylines compliant with the ASAM OpenLabel standard\footnote{\url{https://www.asam.net/standards/detail/openlabel/}, accessed on 18 May 2023.}.

Thus, the main contributions of this paper are:
\begin{itemize}
\item A methodology to provide 3D curbs' detections of a LiDAR point cloud sequence in a standardized output format for being used in an annotation tool.
\item A scan-level curb detector that works on 2D bird's eye view (BEV) images obtained from LiDAR point clouds.
\item A post-processing methodology that transforms the scan-level curb detections into sequence-level three-dimensional polylines. 
\item Validation of the proposed methodology to reduce the annotation time required by a human annotator to obtain curb ground-truth data by $50.99\%$.
\end{itemize}

\section{Related work}

A significant number of works have been published in recent years on the detection of curbs \cite{romero2021road}. Most of these works use the data from the LiDAR to detect curbs\cite{suleymanov2019online,sun20193d,wang2019point,zhao2012curb,rato2021LiDAR}, given the high accuracy in distance measurement this sensor provides \cite{jung2021uncertainty}.
There are two main types of methods for curb detection on LiDAR point clouds: traditional algorithms and DNN-based. Works using traditional detection methods rely on hand-crafted filters and algorithms\cite{baek2020curbscan,sun20193d,chen2015velodyne}. Such methods have been more widely employed in advanced driving systems mainly due to easier deployment in onboard computers. Nevertheless, these approaches have considerable shortcomings when dealing with complex scenarios such as roundabouts or cross-road\cite{bai2022build}. In this context, DNNs have surpassed traditional methods\cite{bai2022build}. 

Among the DNN-based methods, the authors in \cite{suleymanov2019online} use a U-Net network \cite{ronneberger2015u} applied to LiDAR data for visible curb detection and a multilayer convolutional network for partially hidden curb inference. A method based on two stages is proposed in \cite{jung2021uncertainty}, a DNN for visible curb detection and uncertainty quantification. These works focus on real-time scan-level curb detection models and do not exploit the temporal information of the point clouds. In contrast, we propose using a DNN as an initial stage and improving detections by inferring sequence-level curbs.

Some of the most relevant large-scale open-source automotive datasets, such as KITTI, \cite{Geiger2013IJRR}, SemanticKITTI \cite{behley2019semantickitti} or NuScenes \cite{caesar2020nuscenes} do not include curb annotations. Given the lack and need for annotated data in this context, some works focus on offline curbs annotation. In \cite{bai2022build}, authors propose a curb labelling method to extend the SemanticKITTI dataset with curb annotations. However, relying on automatic detections as perfect ground truth is usually too optimistic. The industry often sees manual annotation refinement as mandatory to fix errors \cite{sager2021survey}. Common sources of errors are the domain gap when trained with data from a source distribution and applied to unseen data from a new domain or the incorrect or noisy annotations of training data. When different DNNs are employed subsequently\cite{bai2022build}, errors can accumulate and penalize the final accuracy. Our proposal leverages the scan-level detections inferred by a DNN to post-process the estimated curbs from a sequence-level perspective and generate appropriate polyline annotations in a standardized annotation file.

\begin{figure}[b]
\includegraphics[width=\columnwidth]{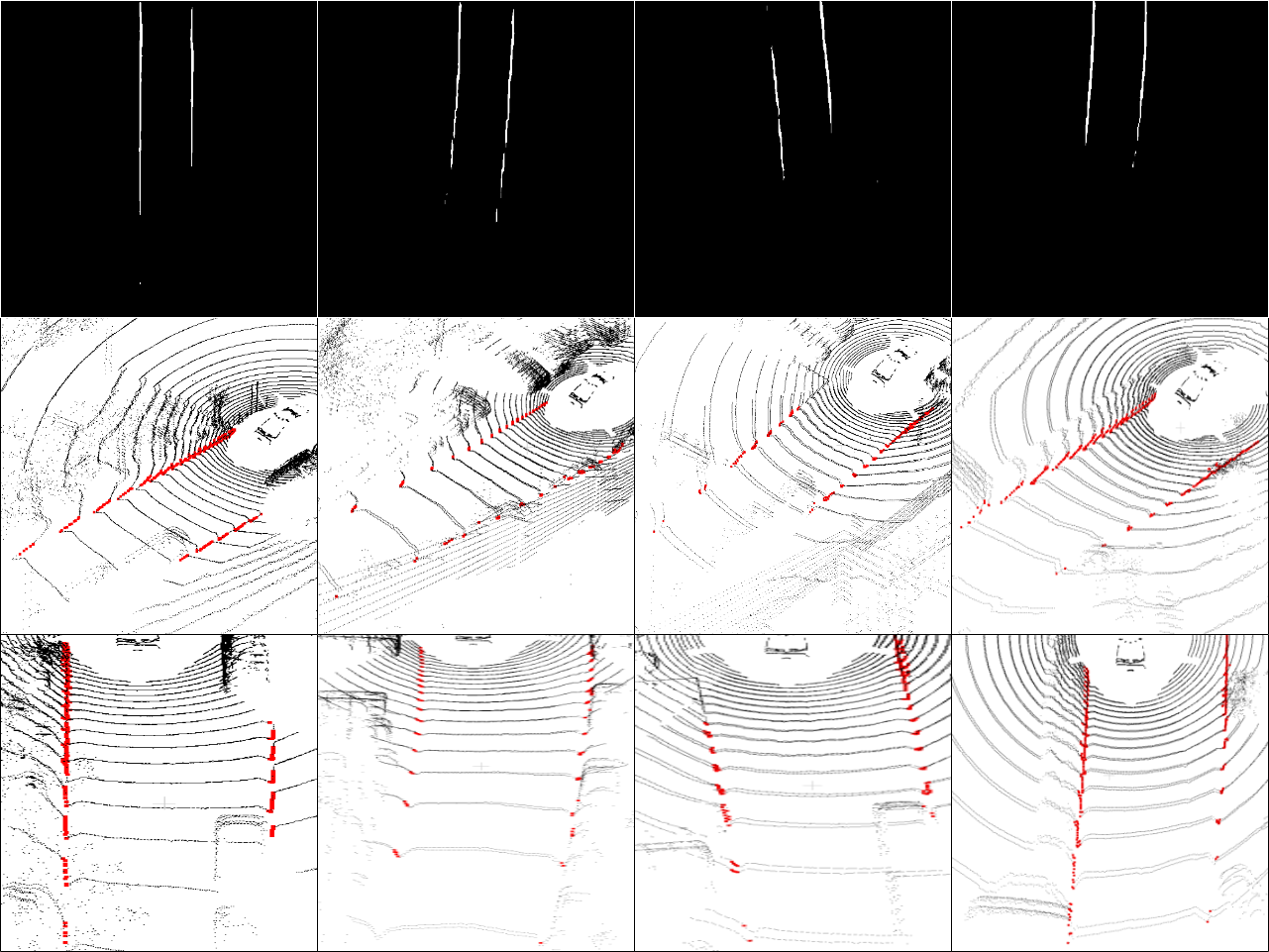}
\caption{Four examples of 2D to 3D detection and transformation. The top row shows DNN inference, i.e. 2D curb detection. In the middle row, in red, the 2D-3D transformation of the detection on the point cloud of the original scan is shown. The bottom row is a top view of the previous one.}
\label{fig_curb_over_pcd_examples}
\end{figure}

\section{Methodology} \label{Methodology}
Our method consists of three main stages, as shown in Fig.~\ref{fig_diag_general_v3}. First, we process point clouds to obtain BEV point cloud representations of the different scans of a sequence. Second, we use a DNN to infer the curbs of each scan. Third, we apply a sequence-level processing step to get 3D curb estimations. This last processing consists of 1) obtaining the 3D points corresponding to the 2D curb detection of each scan, 2) reconstruction of the detected 3D curb points of the whole sequence from the scans of the sequence, the 2D curb detections and the odometry, 3) clustering, skeletonization and simplification of the 3D curbs to generate final polylines in the standardized annotation file.

\subsection{Point cloud to BEV}

To detect curbs, we can use two types of input representation: 3D or BEV projection of the LiDAR point cloud. The 3D option is more accurate and has more information, but it is more complex and computationally expensive
. Therefore, we choose the BEV option as a more compact representation \cite{chen2017multi}.


The point cloud, which is a set of 3D points in space, is divided into M slices corresponding to different height intervals and projected onto a 2D grid map with a specific cell size. Each projection produces a separate height map by encoding the height of the highest point in each grid cell. The BEV is thus encoded as a set of M-channel features.

\subsection{Scan-level curb segmentation}
We propose a semantic segmentation DNN to estimate 2D curbs. This network takes the M-channel BEV maps and infers a pixel-wise 2D mask, where each pixel is assigned a class label, in our case, "curb" and "non-curb". We use the pixel-wise cross-entropy loss to train the network, the most commonly used for semantic segmentation tasks \cite{xu2023comparative}. The loss function is summed for all pixels of the input tensors as follows:

\begin{equation}
\mathcal{L} = - \frac{1}{N} \sum_{i=1}^{N} y_i \log(p_i) + (1 - y_i) \log(1 - p_i)\label{eq}
\end{equation}

where $y_i$ is a binary class label representing the ground truth for pixel $i$, $p_i$ is the softmax predicted class for each pixel, and $N$ is the total number of pixels.


The DNN model allows us to obtain an initial approximation of the curbs to generate the corresponding ground truth. Then, we perform a series of post-processing steps to get more robust and consistent results considering the whole sequence.

\subsection{3D curb refinement}
\textbf{2D to 3D Transformation.} In this stage, the first step is to transform the 2D inferences from the DNN output for each scan into curb 3D points. For this purpose, as shown in Fig.~\ref{fig_diag_general_v3}, we use the LiDAR input scans from which we extract the necessary information to assign the heights to the points classified as curbs in the previous inference. Since the conversion step from point clouds to BEV produces a loss of information generated by the grid resolution and the number of M slices, this transformation is crucial to get a good approximation of the height of the detected curb points (see Fig.~\ref{fig_curb_over_pcd_examples}). To achieve this, we first convert the point cloud's coordinates to pixel positions, keeping the height and the ID of each point ($x_{pixel}$, $y_{pixel}$, $id_{point}$ and $z_{point}$), and extract curbs coordinates from 2D prediction labels ($x_{label}$, $y_{label}$). Then, we merge both data frames joining on x and y coordinates to get a new data frame ($df$) where we assign for each pair of label coordinates ($x_{label}$, $y_{label}$), the corresponding ($id_{point}$, $z_{point}$) which may be one or more. Next, after dropping the $id_{point}$ from the data frame, we select only the records with the minimum $z_{point}$ value. Then, we filter by the height ($z_{point}<0.14m$) \cite{romero2021road}, keeping the points most likely to belong to a curb while adding the $id_{point}$ again by merging with the first data frame ($df$). Now, we have an $id_{point}$ for each pixel detected as curb ($x_{pixel}$, $y_{pixel}$) and the most probable height ($z_{point}$). Taking these $id_{point}$, we have the curbs' 3D points for each scan.


\textbf{Reconstruction.} In this step, all the curbs' 3D points of each scan (obtained in the previous step), together with the input point clouds, are considered. Applying the odometry information, a cumulative reconstruction of all of them is made to obtain a curb point cloud of the entire sequence.

\textbf{Annotation Generation.} At this point, we first group the detections in different curbs so that in later stages, we can manipulate them separately, for example, by loading them effectively in the labelling tools or deleting those sections that correspond to false detections. To carry out this separation, we perform a \textbf{\textit{clustering}} stage using the DBSCAN algorithm \cite{ester1996density}. This algorithm is appropriate for clusters that present a similar density in their data, so before its application, we perform a voxel subsampling process that balances the densities of the clusters, which is also helpful for the following interpolation step.

To obtain the final polyline curb representation, we use the \textbf{\textit{skeletonization}} algorithm \cite{bucksch2009skeltre}. By applying this algorithm, specially designed for the skeletonization of a point cloud obtained from a LiDAR, we obtain the linear traces we expect to represent the curbs. 

The last post-processing step is the \textbf{\textit{simplification}}. In this step, we reduce the number of points of the detected curb skeletons by means of the Ramer-Douglas-Peucker algorithm \cite{ramer1972iterative,douglas1973algorithms}. This algorithm uses a given distance tolerance to determine which points on a line are to be eliminated or retained. 

\begin{figure}
\includegraphics[width=\columnwidth]{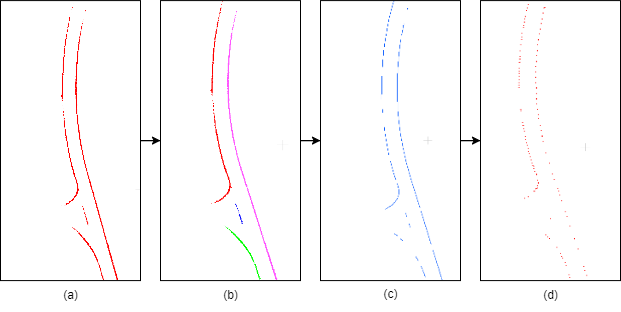}
\caption{Post-processing pipeline. (a) Reconstruction. (b) Clustering. (c) Skeletonization. (d) Simplification.}
\label{fig_postprocessing}
\end{figure}

A representation of the reconstruction and the subsequent post-processing steps (until the simplified point cloud is obtained) can be seen in Fig.~\ref{fig_postprocessing}.

\textbf{Annotation File}. In this step, the curbs are stored as polylines in files that follow the ASAM OpenLabel standard.

\begin{figure*}
\centering
\includegraphics[width=\textwidth]{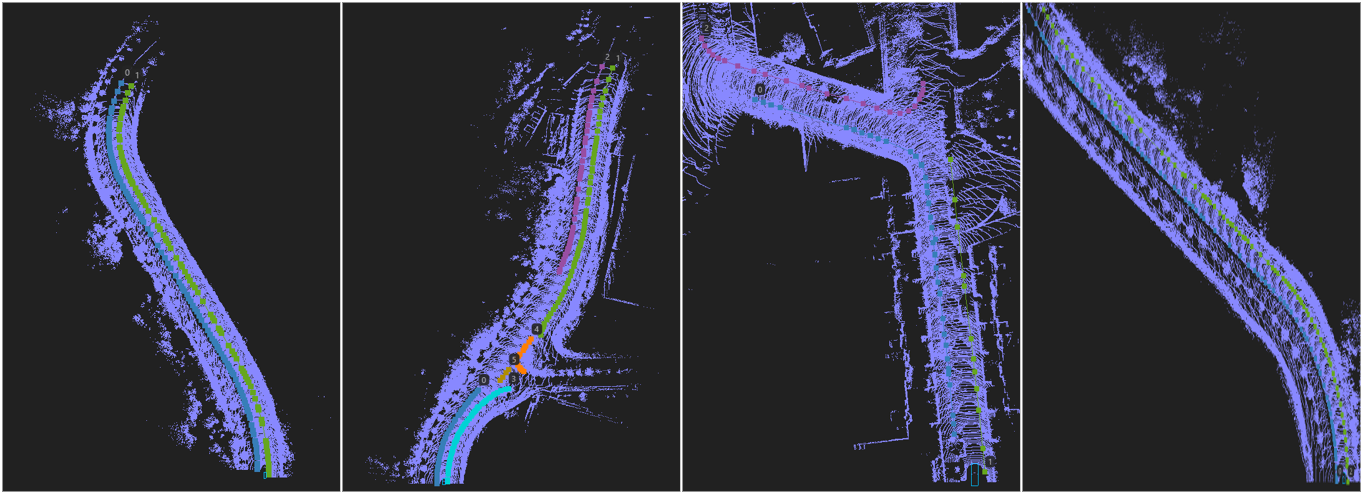}
\caption{Four different point cloud reconstruction maps (not on the same scale) we used in the annotation experiments, along with their corresponding ground truths (labelled with numbers).}
\label{fig_maps_gt}
\end{figure*}

\section{Experimental Results}

\subsection{Implementation details}
\textbf{DNN Training Datasets.} To train, validate, and test the neural network, we use two public datasets \cite{jung2021uncertainty,bai2022build}. The first dataset contains two groups of data: a first group of 5224 frames with curb labels in line format and a second group of 6820 frames with labels in point format. Both groups have BEV representations of 3D point clouds with a $0.1m/pixel$ resolution. 
The first group gave imprecise estimations due to the wide shape of lines in the labels. Therefore, we used the second group with labels in point format and with a larger number of frames. However, the labels were converted from point format to line format using a single pixel width to avoid the problems encountered with the first group. The BEV representations also have a resolution of $0.1m/pixel$.



The second dataset is based on the public LiDAR semantic dataset SemanticKITTI \cite{behley2019semantickitti}. It consists of curb annotations for the first 11 sequences from KITTI. The KITTI segmentation point clouds and the ground truths provided by \cite{bai2022build} are adapted in the same way as the previous dataset, transforming the point clouds to BEVs based on the height (6 slices) and the annotations to one-pixel line ground truths. This second dataset consists of 23101 frames, which, added to the previous 6820, gives us a total of 29921 frames for the overall dataset.

\textbf{Dataset Distribution.} To make the distribution of the frames available in the database regarding training, validation, and testing, we consider the previous divisions that already existed in \cite{jung2021uncertainty,bai2022build}. In the case of \cite{jung2021uncertainty}, there is a block of 2000 separate frames for testing and the rest, 4820, for training. We maintain this division by adding a second one in the test block, separating (randomly) 1500 for validation and 500 for the test. Of the 11 KITTI sequences used in \cite{bai2022build}, we separate 3 for validation (03, 06, 07) and 1 for testing (01), obtaining 3003 and 1101 frames, respectively. The separation between validation and testing aims to maintain a certain distance between the scans used for training on the one hand and validation and testing on the other. In this way, we can better evaluate how our models generalize. The final division of the dataset is as follows: 23917 for train, 4503 for validation, and 1601 for testing. This produces a distribution of approximately $80\%$ of the dataset dedicated to training, $15\%$ to validation, and $5\%$ to testing.


\textbf{DNN Training and Testing.} We use a semantic segmentation network to identify and separate different curb instances in an image (i.e., assigning a label to each pixel). Specifically, we adopt the DNN DeepLabv3+ \cite{chen2018encoder}. DeepLabv3+ is a state-of-the-art model for image segmentation tasks. It is based on fully convolutional networks (FCN), allowing efficient and accurate pixel-level predictions. A different segmentation network could also be used in our pipeline. We use DeepLabv3+ with a ResNet-50 \cite{he2016deep} backbone and initialized with pre-trained weights on the ImageNet dataset \cite{imagenet_cvpr09}. 

To further improve the variability of the scenes used for DNN training, we apply a data augmentation random process \cite{info11020125}. For training our DeepLabv3+ network, we use an experimental setup composed of an NVIDIA GPU GV100GL graphic card, Ubuntu 18.04 operating system, and the TensorFlow framework. The training images, with a size of $512 \times 512$ pixels, are fed to the network in batches of 16 frames and a learning rate of 0.0001 during a maximum of 50 epochs. An Adam optimizer and the pixel-wise cross-entropy loss complete the configuration parameters. For testing, we use the subset of 1601 test images. As metrics, for a pixel-based evaluation in the BEV space, we follow the criteria employed in the KITTI-ROAD dataset \cite{fritsch2013new}. We measure the precision, recall, and F-score.


\subsection{Experiments}
This section is divided into two parts, one in which we present the results of the DNN to infer scan-wise detections using the BEV representations and another dedicated to the usage of our estimated 3D curbs for ground truth generation.


\textbf{Single-scan curb estimation.}
Table \ref{table:1} shows the results of the segmentation network for two different tolerances, considering that 1 pixel corresponds to $0.1m$, according to the spatial resolution considered.

\begin{table}[htbp]
\caption{Precision, Recall, and F-score of the DNN on the test set.}
\label{table:1}
\begin{center}
\begin{tabular}{c c c c }
\hline
\textbf{Tolerance} & \textbf{Precision} & \textbf{Recall} & \textbf{F-score}\\
\hline
\hline
1px & 0.738 & 0.673 & 0.661  \\
\hline
3px & 0.907 & 0.828 & 0.814  \\
\hline
\end{tabular}
\label{tab1}
\end{center}
\end{table}

The F-score obtained with a tolerance of 3 pixels shows that the curbs are detected in most cases and the suitability of the DNN for a first curb estimation. When the tolerance is more restrictive, the metrics decrease slightly, which motivates the need for the second refinement stage in our pipeline.

\begin{figure}[b]
\includegraphics[width=\columnwidth]{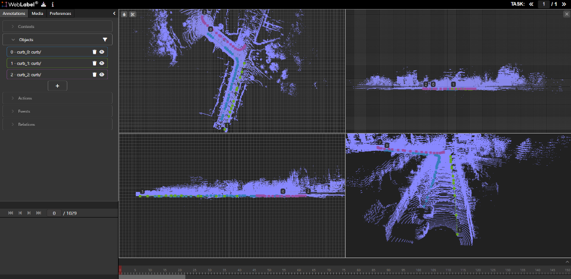}
\caption{Example of curb annotation in our labelling tool showing four viewpoints of a map. Three curbs’ ground truth are annotated as coloured polylines.}
\label{fig_weblabel_4windows}
\end{figure}

\textbf{3D curb annotation.}
We conducted a series of tests to evaluate the usefulness of the curb estimations obtained by our method for ground truth generation. These tests involved annotating curbs with and without pre-annotations obtained with our method. Both annotation tasks were carried out by four non-expert annotators who were familiar with the used annotation tool \cite{mujika2019web}. 
The tests were performed on four sequences, collected from a LiDAR sensor installed on a prototype vehicle, that contained different curb characteristics such as straight sections, curves, obscured by vegetation, parked vehicles, etc. (see Fig.~\ref{fig_maps_gt}). An example of the annotation tool used with labelled curbs can be seen in Fig.~\ref{fig_weblabel_4windows}. 
Due to the sparse and low-resolution nature of LiDAR point clouds, the process of manual annotation in general, and of curbs in particular, is a complex and tedious task that requires a certain amount of skill on the part of the human annotator. To facilitate the annotation process, annotators are instructed to use the top view coloured with a gradient on the z-axis that highlights the discontinuities in the height of the curbs.

We have used accurately crafted manual ground truths to evaluate the annotations made by the annotators. In the annotation process, the annotators were asked to measure the annotation times for each map, both for the ones annotated from scratch and with pre-annotations. 
The evaluation metrics employed (recall, precision, and F-score) have been obtained using the process presented in \cite{martirena2020automated}. This approach converts sets of 3D polylines into sets of 3D points and samples the polylines using a specific metric step size. Subsequently, it compares these 3D polylines by utilizing 3D Euclidean distances. It is important to note that the maximum error in distance measurement is restricted to half the discretization step. We have considered the resolution of $0.1 m/pixel$ imposed by the conversion mentioned earlier in the implementation details.

In Table \ref{table:2}, the average values of the metrics obtained for the annotations made without and with pre-annotations of each map are shown, as well as the global average and the improvement achieved thanks to the use of pre-annotations. Based on the BEV resolution, a tolerance of $10cm$ was used to count the annotations as correct.
Although there are no major differences between the values obtained without pre-annotations and with pre-annotations, mainly because human annotators ultimately make all the annotations, we do notice a slight improvement in those made from the pre-annotations generated by our method. %
The minor time improvement obtained for map 3 deserves special mention due to the errors both in annotating parts that are not real curbs (false positives) and curbs that were not identified by the annotator or model (false negatives). We believe that in some scenarios, it might be difficult to correctly annotate curbs with no further information (e.g., from an RGB camera).
Also, we show in Table \ref{table:3} that using pre-annotations provided by our method reduces the annotation time by $50.99\%$. The effect of pre-annotations on the improvement in annotation times can be explained if we take as a reference the 90.7\% precision obtained for a tolerance of 3 pixels. This means that almost all pre-annotations do not need correction, just a visual check that requires little time. The annotator would only have to annotate the rest of the unannotated curbs (the missing part in the recall to reach 100\%, i.e. 17.2\%), and correct the wrong annotations, i.e. 9.3\%. The annotation time should be dedicated to the visual verification of the pre-annotations (82.8\%), annotation of the undetected ones (17.2\%), and correction of the bad detections (9.3\%).


\begin{table}[t]
\caption{Precision, Recall and F-score of the annotations tests }
\label{table:2}
\begin{center}
\begin{tabular}{ccccc}
\hline
\multicolumn{1}{l}{} & \textbf{Map} & \textbf{Recall} & \textbf{Precision} & \textbf{F-score} \\
\hline
\hline
\multirow{5}{*}{Without   pre-annotations} & 1 & 0.919 & 0.919 & 0.919 \\
 & 2 & 0.890 & 0.878 & 0.883 \\
 & 3 & 0.617 & 0.687 & 0.650 \\
 & 4 & 0.940 & 0.934 & 0.937 \\
 & \textbf{Average} & \textbf{0.841} & \textbf{0.855} & \textbf{0.847} \\
\hline
\multirow{5}{*}{With   pre-annotations} & 1 & 0.905 & 0.913 & 0.908 \\
 & 2 & 0.889 & 0.895 & 0.892 \\
 & 3 & 0.673 & 0.719 & 0.695 \\
 & 4 & 0.980 & 0.985 & 0.983 \\
 & \textbf{Average} & \textbf{0.862} & \textbf{0.878} & \textbf{0.870} \\
\hline
\hline
\textbf{Average Improvement} & \textbf{\%} & \textbf{2.408} & \textbf{2.750} & \textbf{2.647} \\
\hline
\end{tabular}
\label{tab2}
\end{center}
\end{table}

\begin{table}[t]
\caption{Improvement in annotation times (hh:mm:ss)}
\label{table:3}
\begin{center}
\begin{tabular}{c c c c }
\hline
\multicolumn{1}{c}{\textbf{Map}} & \multicolumn{1}{c}{\textbf{\begin{tabular}[c]{@{}c@{}}Average Time  \\ without \\ pre-annotations \end{tabular}}} & \multicolumn{1}{c}{\textbf{\begin{tabular}[c]{@{}c@{}}Average Time  \\ with \\ pre-annotations \end{tabular}}} & \multicolumn{1}{c}{\textbf{\begin{tabular}[c]{@{}c@{}}Improvement \\ (\%)\end{tabular}}}  \\
\hline
\hline
1 & 00:41:41 & 00:14:52 & 64.33 \\ 
\hline
2 & 00:34:52 & 00:16:14 & 53.47 \\ 
\hline
3 & 00:10:53 & 00:10:23 & 4.60 \\ 
\hline
4 & 00:17:18 & 00:09:52 & 43.02 \\ 
\hline
\hline
Total & 01:44:44 & 00:51:20 & \textbf{50.99} \\ 
\hline
\end{tabular}
\label{tab3}
\end{center}
\end{table}

\addtolength{\textheight}{0cm}   




\section{Conclusions}

The need for ground truth data in AD tasks requires large amounts of data that need to be manually labelled. Semi-automatic annotation algorithms can help reduce human annotation time, which results in cost savings during the annotation process. 
In this paper, we present an approach to generate 3D curb pre-annotations in the ASAM OpenLABEL standardised output format from a sequence of LiDAR point clouds. 
Our method detects curbs at a scan level and refines them in a second sequence-level post-processing stage. Final detections are stored as polylines.
By performing a manual annotation campaign with real data obtained from a LiDAR sensor equipped in a testing vehicle, we have validated the suitability of our method. Our proposed curb detection pipeline reduces manual annotation time by $50\%$ while keeping a similar accuracy in the annotations. Future work involves training the DNN with a larger dataset and exploring alternative architectures to enhance the pre-annotations. We also plan to extend the method to incorporate data from an RGB camera to enhance the detections' reliability in the most difficult scenarios.


\section*{Acknowledgment}
This work has received funding from Basque Government under project AutoEv@l of the program ELKARTEK 2021.

\bibliographystyle{ieeetr}
\bibliography{curbs_detection.bib}

\end{document}